\documentclass[journal]{IEEEtran}
\usepackage{amsmath,amssymb,amsfonts,bm}
\usepackage{booktabs}
\usepackage{multirow}
\usepackage{algorithm}
\usepackage{algorithmic}
\usepackage{graphicx}
\graphicspath{{paper/}}
\usepackage{cite}
\usepackage{url}
\usepackage{tikz}
\usetikzlibrary{arrows.meta,positioning,calc}

\title{Clean-Reference Streaming Detection of Lens Occlusion and Photometric Transitions for Camera Tamper Monitoring}

\author{Bo~Ma,~Wei~Qi~Yan,~and~Jinsong~Wu%
\thanks{Bo Ma and Wei Qi Yan are with Auckland University of Technology, Auckland 1024, New Zealand.}%
\thanks{Jinsong Wu is with Guilin University of Electronic Technology, Guilin, China.}%
\thanks{Corresponding author: Bo Ma (e-mail: rcn4743@aut.ac.nz).}%
}

\begin{document}
\maketitle

\begin{abstract}
A surveillance camera is an image sensor whose silent physical degradation invalidates every downstream consumer of its data. In-situ integrity alarms for such vision sensors require low false-alarm rates, bounded computation, and diagnosable behavior under nuisance illumination changes. This paper studies a deliberately narrow streaming integrity monitor for two low-cost sensor-fault signatures: texture-collapsing lens occlusion and abrupt photometric scene transition. The detector compares sampled luminance and local-gradient statistics with a clean-only sliding reference, applies coarse-grid structured-light rejection and mode/rapid-brightness suppression, and emits at most one notification per tamper episode. We formalize the decision predicates and derive a consistency rule for when rapid-brightness suppression makes the scene-transition path unreachable. On 320 in-scope controlled sequences, the default state machine attains 0.800 F1 and 0.822 balanced accuracy (significantly better paired correctness than the strongest baseline, though the F1 margin is not statistically resolved); on a magnitude-swept public audit it attains the highest partial AUC under a 5\% false-alarm budget, and a separate extended-stress FPR-constrained sweep reaches 0.925 recall at 0.025 false-positive rate. Public Xiph, Bremen IoT, and UHCTD diagnostics show the fixed predicates preserve low false alarms while recall concentrates inside the declared envelope (UHCTD in-scope covered recall 0.667 versus 0.016 out of scope), and a 9.09-camera-hour verified-negative public audit records zero false alarms. The method is best interpreted as an auditable sensor-health subsystem rather than a universal camera-tamper classifier.
\end{abstract}


\section{Introduction}
A surveillance camera is, above all, an image sensor, and physical interference with that sensor --- covering the lens, changing its focus, moving the camera --- silently invalidates all analytics downstream of image acquisition, while day--night switching, exposure control, room-light toggles, and large foreground objects produce equally abrupt image changes. This is a sensor-reliability problem: a single camera-health signal gates the availability and trustworthiness of every downstream consumer of the sensor's data, so a useful in-situ integrity monitor must balance response latency, nuisance rejection, predictable computational cost, and diagnosable behavior. Throughout the paper, \emph{tamper} names the physical event class, while \emph{sensor-integrity} or \emph{camera-health} names the system-facing signal that reports it.

Prior systems compare current frames with buffered references using histograms, edges, frequency coefficients, keypoints, or learned representations \cite{ribnick2006realtime,saglam2009adaptive,lee2015unified,mantini2017signal}. The present work addresses a narrower engineering point: how low-cost photometric and texture cues can be composed into a stateful decision rule whose baseline cannot absorb an active alarm and whose suppression logic is auditable. The novelty is this stateful, auditable composition rather than the individual scalar cues: an active alarm cannot update the clean reference, nuisance transitions are handled by explicit negative gates, and the emission policy is episode-based rather than per-frame --- a trade that matters in embedded pipelines where deterministic behavior, bounded state, and failure diagnosis outweigh an opaque classifier.

The method is deliberately not presented as a universal covered--defocused--moved camera classifier. Its positive scope comprises (i) lens obstruction producing a sufficiently large texture collapse with a bounded luminance change and (ii) a scene transition producing a sufficiently large luminance change; pure rotation or translation that preserves global statistics, subtle local occlusion, defocus without the required mean shift, and video-file forgery all fall outside that positive guarantee.

The contributions are:
\begin{itemize}
\item a clean-reference streaming state machine for two declared vision-sensor fault signatures: texture-collapsing lens occlusion and abrupt photometric scene transition;
\item an implementation-faithful formulation of the mean, gradient, grid, suppression, and episode predicates, with diagnostic metadata for field triage;
\item a configuration-safety condition showing when rapid-brightness suppression makes the scene-transition path unreachable;
\item a false-alarm-constrained evaluation protocol with paired baselines, targeted ablations, and public-data diagnostics; and
\item a deployment boundary showing where fixed photometric predicates end and registration, calibration, or learned references begin.
\end{itemize}

\begin{figure*}[!t]
\centering
\begin{tikzpicture}[
  font=\footnotesize,
  block/.style={draw, rounded corners=1pt, align=center, inner sep=3pt,
                minimum height=8mm},
  gate/.style={draw, dashed, rounded corners=1pt, align=center, inner sep=3pt},
  state/.style={draw, rounded corners=2pt, align=center, inner sep=3pt,
                minimum height=8mm, fill=black!5},
  flow/.style={-{Stealth[length=2mm]}, thick},
  soft/.style={-{Stealth[length=2mm]}, dashed}
]
\node[block] (input) at (0,0) {Frame $I_t$};
\node[block, text width=24mm] (stats) at (3.6,0)
  {Sampled statistics\\ $\mu_t$, $g_t$, $32{\times}32$ grid};
\node[block, text width=26mm] (ref) at (3.6,2.4)
  {Clean-only reference\\ $(\mu_t^\star, g_t^\star,\ \mathrm{grid}^\star)$};
\node[block, text width=30mm] (scene) at (8.2,1.1)
  {Scene path\\ $\delta_t>\theta_{\mathrm{chg}}$, Eq.~\eqref{eq:scene}};
\node[block, text width=30mm] (occ) at (8.2,-1.1)
  {Occlusion path\\ texture collapse, Eq.~\eqref{eq:occlusion}};
\node[gate, text width=26mm] (modegate) at (8.2,2.9)
  {Mode and rapid-brightness gates};
\node[gate, text width=26mm] (gridgate) at (8.2,-2.9)
  {Coarse-grid gate,\\ Eq.~\eqref{eq:coarse}};
\node[state, text width=26mm] (episode) at (12.6,0)
  {Episode state:\\ one notification\\ per episode};
\node[block] (output) at (16.2,0)
  {$\mathbf{y}_t=(d_t, c_t, z_t, m_t)$};
\node[state, text width=26mm] (clean) at (12.6,-2.9)
  {Clean state:\\ reference updates\\ resume};

\coordinate (split) at ($(stats.east)+(6mm,0)$);
\draw[flow] (input) -- (stats);
\draw[flow] (stats.east) -- (split);
\draw[flow] (split) |- (scene.west);
\draw[flow] (split) |- (occ.west);
\draw[soft] (stats) -- node[right, inner sep=2pt] {clean frames only} (ref);
\draw[flow] (scene.east) -- ($(scene.east)+(5mm,0)$)
  |- ($(episode.west)+(0,2.5mm)$);
\draw[flow] (occ.east) -- ($(occ.east)+(5mm,0)$)
  |- ($(episode.west)+(0,-2.5mm)$);
\draw[soft] (modegate) -- node[right, inner sep=2pt] {suppress} (scene);
\draw[soft] (gridgate) -- node[right, inner sep=2pt] {reject} (occ);
\draw[flow] (episode) -- (output);
\draw[flow] (episode.south) -- node[right, inner sep=2pt]
  {clean recovery frame} (clean.north);
\draw[soft] (clean.south) -- ($(clean.south)+(0,-6mm)$)
  -| node[below, pos=0.25, inner sep=3pt] {reference updates resume}
  (stats.south);
\end{tikzpicture}
\caption{Streaming state-machine view of the detector. Sampled statistics feed
the photometric scene path and the gradient-collapse occlusion path; the mode
and rapid-brightness gates suppress nuisance transitions, and the coarse-grid
gate rejects structured lighting; a clean recovery frame ends the
single-notification episode and resumes clean-only reference updates.}
\label{fig:pipeline_state}
\end{figure*}
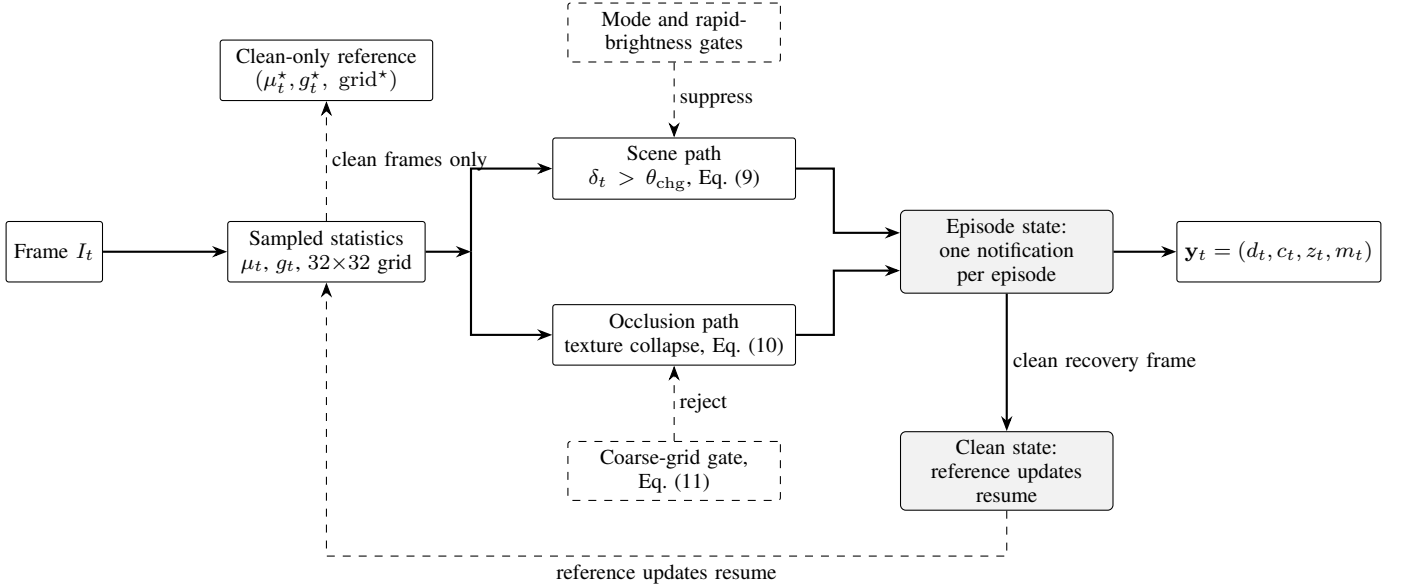

\section{Related Work}
\subsection{Reference-Based and Hand-Crafted Detectors}
Ribnick \emph{et al.} compared buffered old and recent frames using normalized distance measures \cite{ribnick2006realtime}; Saglam and Temizel developed adaptive detectors for obstruction, displacement, and focus changes emphasizing low false-alarm operation \cite{saglam2009adaptive}; Tung \emph{et al.} classified tamper events over an adaptive background codebook \cite{tung2012codebook}; and Roy \emph{et al.} proposed a robust background-model update that resists a slowly corrupted reference \cite{roy2024rsvddpd}. These methods establish the value of a temporal reference but also expose reference contamination and threshold calibration as practical concerns. Edge loss is a particularly common cue because cover and defocus remove high-frequency content: Lee \emph{et al.} measured edge disappearance relative to a background image with foreground-object exclusion \cite{lee2015unified}, Shi \emph{et al.} combined background construction with local-binary-pattern similarity \cite{shi2021occlusion}, and Sitara and Mehtre added foreground coverage area and its temporal stability \cite{sitara2018sabotage}.

\subsection{Camera Displacement and Registration-Based Detection}
Because global mean and sampled gradient are largely insensitive to a rigid viewpoint change, several prior systems add an explicit geometric cue, from SIFT keypoint-geometry verification \cite{yi2013sift} to the registration and motion-compensation strategies surveyed by Chapel and Bouwmans \cite{chapel2020movingcamera}. We do not implement a registration baseline in this paper; Section~\ref{sec:iot_crosscheck} instead measures, on real camera-disjoint images, how far a purely photometric predicate is from covering this case on its own.

\subsection{Statistical, Learned, and Calibrated References}
Mantini and Shah modeled normal signal activity with a Gaussian mixture and temporal filtering \cite{mantini2017signal}, later learned a generative reference with deep feature comparison \cite{mantini2019generative}, released the UHCTD dataset with deep-classifier benchmarks on covered/defocused/moved/normal frames --- the closest prior work to a learned camera-tamper baseline \cite{mantini2019uhctd} --- and surveyed which hand-crafted feature families transfer best across cameras \cite{mantini2023survey}. Pan detected physical camera-integrity attacks with a deep video-frame-interpolation anomaly score \cite{pan2019interpolation}. Such learned models represent broader appearance variation but introduce training data, model storage, and domain-transfer requirements; Sidnev \emph{et al.} highlighted the difficulty of per-camera threshold selection even for low-level features \cite{sidnev2018calibration}. More broadly, camera tamper detection is one instance of video anomaly detection \cite{liu2024generalized}, and false-alarm-constrained evaluation is standard in adjacent visual-inspection settings \cite{bao2022miad}; our evaluation adopts the same convention.

Our approach occupies a different point in this design space: no pixelwise background modeling or foreground masks \cite{lee2015unified}, no trained probabilistic or deep reference \cite{mantini2017signal,mantini2019generative,mantini2019uhctd}, no automatic calibration claim \cite{sidnev2018calibration}, and no geometric registration cue \cite{yi2013sift,chapel2020movingcamera}. The distinctive element is the composition of sampled statistics, clean-only reference maintenance, structural nuisance rejection, and explicit operational state; the intended advantages are simplicity, auditability, and bounded state, with geometric invariance and cross-camera tuning delegated to complementary modules.

\subsection{Embedded Analytics and Sensor Self-Validation}
Hand-crafted cues remain relevant when throughput and resource bounds dominate, as in FPGA \cite{kryjak2012fpga}, DSP \cite{lin2012dsp}, low-power smart-camera \cite{gaibotti2015lowpower}, and single-board-computer \cite{adamovskiy2026smoke} tamper detectors; this body of work motivates our emphasis on bounded, auditable arithmetic evaluated as a hardware-agnostic algorithmic contribution. Camera tampering is also an instance of the broader sensor-fault-detection problem: self-validating sensor concepts attach an online validity status and diagnostics to every measurement \cite{henry1993seva}, surveys and classifiers for outlier and fault detection in wireless-sensor and IoT deployments organize the corresponding principles for low-cost distributed sensors \cite{zhang2010outlier,gaddam2020iotfaults,zidi2018svmfault}, and automotive perception likewise treats camera degradation as a sensor-level reliability concern \cite{pandharipande2023automotive}. The present detector follows the same design logic at the level of an image sensor --- validate the stream in situ with bounded arithmetic, attach diagnostic metadata to every alarm, and expose a health signal that gates downstream analytics --- motivating auditability and false-alarm control over maximal classification power.

\section{Problem Scope and Output Semantics}
Let $I_t\in[0,255]^{H\times W}$ be a grayscale frame acquired at timestamp $\tau_t$. The detector emits $\mathbf{y}_t=(d_t,c_t,z_t,m_t)$, where $d_t\in\{0,1\}$ is an alarm, $c_t\in[0,1]$ is a cue magnitude rather than a calibrated posterior probability, $m_t$ is diagnostic metadata, and $z_t\in\{\mathrm{no\ tamper},$ $\mathrm{occlusion\ event},$ $\mathrm{scene\ transition}\}$. The second label denotes an abrupt \emph{photometric} scene transition. It should not be read as a registration-based guarantee for every camera displacement.

The detector holds a window of at most $N$ frames accepted as clean. Its scalar reference is $(\mu_t^\star,g_t^\star)$, and its structural reference is a per-cell mean on a $32\times32$ grid. Frames declared tampered are never inserted into these references. The declared operating envelope is: \emph{in scope} --- a texture-collapsing cover with a bounded mean shift, and an abrupt photometric scene transition not suppressed by the mode or rapid-brightness gates; \emph{out of scope or weak} --- pure camera rotation or translation (global statistics carry no registration evidence), mean-preserving defocus, and subtle local partial covers; \emph{intended nuisance rejection} --- global illumination steps; \emph{ambiguous, requiring field validation} --- foreground objects close to the lens; and \emph{excluded from accuracy metrics} --- unreadable or failed clips (each condition is tabulated with its mechanistic reason in the supplementary material).

\subsection{Clip-Level Ground Truth for Evaluation}
\label{sec:ground-truth-definition}
The per-frame alarm $d_t$ is a \emph{prediction}. Evaluating that prediction requires a separate, independently defined \emph{ground-truth} label, and the two must not be conflated. We define them as follows.

A clip is ground-truth \emph{positive} if and only if at least one declared in-scope condition genuinely occurred at some point during the \emph{entire} recording, independent of whether the detector raised an alarm, and ground-truth \emph{negative} only if no such condition occurred anywhere in the full recording. This is a claim about the physical recording session, not a sampled subset of its frames; a brief or previously unnoticed occurrence anywhere in the clip makes it positive.

Consequently, ``the detector did not alarm'' is not evidence that a clip is ground-truth-negative: a predicted-negative clip with a missed genuine event is a false negative, and a true negative requires independent confirmation --- exhaustive review of the full recording or an authoritative collection record, not a coarse pass over sampled frames or the absence of a reported incident. This definition governs every true-negative, false-negative, specificity, and false-positive-rate statement in the paper; where labels do not meet it, we report a conservative accounting convention instead of a true-negative count.

\section{Streaming Tamper Detector}
\subsection{Sampled Photometric and Texture Statistics}
Let $x$ index columns and $y$ index rows throughout, so $x\equiv0\pmod 4$ selects every fourth column. Mean extraction visits every row and every fourth column:
\begin{equation}
\begin{aligned}
\Omega_\mu=\{(x,y):&\ 0\le x<W,\ 0\le y<H,\\
& x\equiv0\pmod 4\}.
\end{aligned}
\end{equation}
The sampled mean is
\begin{equation}
\mu_t=\frac{1}{|\Omega_\mu|}\sum_{(x,y)\in\Omega_\mu} I_t(x,y).
\end{equation}
The local-gradient proxy samples both axes every four pixels but compares immediate neighbors:
\begin{equation}
\begin{aligned}
\Omega_g=\{(x,y):&\ 0\le x<W-1,\ 0\le y<H-1,\\
& x\equiv0\pmod 4,\ y\equiv0\pmod 4\}.
\end{aligned}
\end{equation}
\begin{equation}
g_t=\frac{1}{2|\Omega_g|}\sum_{(x,y)\in\Omega_g}
\left(d^x_t(x,y)+d^y_t(x,y)\right),
\end{equation}
where $d^x_t=|I_t(x+1,y)-I_t(x,y)|$ and
$d^y_t=|I_t(x,y+1)-I_t(x,y)|$.
Immediate differences avoid aliasing a one-pixel texture into an apparently flat image; border pixels lacking a right or lower neighbor are excluded.

During warm-up, a frame is treated as boot-time occlusion if
\begin{equation}
\mu_t<\theta_{\mathrm{boot,dark}}\ \lor\
g_t<\theta_{\mathrm{boot,flat}},
\end{equation}
with implementation defaults $\theta_{\mathrm{boot,dark}}=15$ intensity levels and $\theta_{\mathrm{boot,flat}}=2$ gradient levels (the controlled evaluation of Section~V disables the flatness check).
Such frames are emitted but excluded from warm-up. Recovery clears the partially collected reference before accepting new clean frames. Once $N$ clean frames are available, running sums give constant-time reference updates:
\begin{equation}
\mu_t^\star=\frac{1}{N}\sum_{k\in\mathcal{W}_t}\mu_k, \qquad
g_t^\star=\frac{1}{N}\sum_{k\in\mathcal{W}_t}g_k.
\end{equation}

\subsection{Photometric Scene-Transition Path}
Define the inter-frame mean jump $\delta_t=|\mu_t-\mu_{t-1}|$, current and previous texture-collapse indicators
\begin{equation}
o_t=\mathbb{I}[g_t<\theta_{\mathrm{occ}}g_t^\star],\qquad
o_{t-1}=\mathbb{I}[g_{t-1}<\theta_{\mathrm{occ}}g_t^\star],
\end{equation}
where both frames are deliberately tested against the \emph{current} reference $g_t^\star$: the clean window is frozen while a candidate is active, so $g_t^\star$ is the most recent uncontaminated texture reference for either frame,
and normalized gradient change
\begin{equation}
q_t=\frac{g_{t-1}-g_t}{\max(g_t^\star,\epsilon)}.
\end{equation}
The primary scene candidate is
\begin{equation}
\mathcal{C}^{\mathrm{sc}}_t=
[\delta_t>\theta_{\mathrm{chg}}]\land[\neg o_t\lor o_{t-1}]
\land[\rho_{\mathrm{grad}}\le0\lor q_t\ge\rho_{\mathrm{grad}}].
\label{eq:scene}
\end{equation}
A counter requires $K$ consecutive candidates when persistence is enabled. A secondary luminance route compares the current mean with both the last accepted clean mean and the sliding mean, requiring an absolute jump above $\theta_{\mathrm{chg}}$ and a relative jump above the auxiliary ratio $\rho_{\mathrm{lumin}}$ (default 0.8); in the evaluated pathway it adds baseline-relative luminance memory only.

\subsection{Gradient-Collapse Occlusion Path}
If the scene path has not fired, an occlusion candidate requires the conjunction
\begin{equation}
\begin{aligned}
g_t &<\theta_{\mathrm{occ}}g_t^\star,\\
\Delta_{\min}<|\mu_t-\mu_t^\star|&\le\Delta_{\max},\\
\mu_{\min}\le\mu_t&\le\mu_{\max},
\end{aligned}
\label{eq:occlusion}
\end{equation}
so that Eq.~\eqref{eq:occlusion} denotes all three conditions jointly. The lower bound is $\Delta_{\min}=20$ intensity levels in the implementation; the upper bound and absolute luminance interval reject extreme environmental transitions. Consequently, defocus that reduces texture without also satisfying the mean constraint is not guaranteed to trigger.

For each frame, a $32\times32$ center-sampled luminance grid is compared with its clean reference. Let $r_{20}$ be the fraction of cells changing by more than 20, $\bar\sigma_b$ the mean standard deviation over $4\times4$ grid blocks, and $r_{\mathrm{flat}}$ the fraction of blocks with standard deviation below 3. A low-gradient candidate is rejected as structured lighting change when
\begin{align}
&(r_{20}>0.85\land\bar\sigma_b>10\land r_{\mathrm{flat}}<0.25)\nonumber\\
&\quad\lor(\bar\sigma_b>18\land r_{\mathrm{flat}}<0.20
\land g_t/g_t^\star>0.40).
\label{eq:coarse}
\end{align}
The grid is thus a \emph{negative gate} for occlusion, not a third positive detector.

\subsection{Transition Suppression and Configuration Consistency}
A change in the externally supplied day--night flag (the operating-mode gate, hereafter the mode gate) suppresses scene output for that transition; on typical cameras this flag derives from the IR-cut-filter state or the sensor front end's exposure/gain metadata. Separately, if $\delta_t>\theta_{\mathrm{rapid}}$, the timestamp is stored and scene output is suppressed for $T_{\mathrm{sup}}$. These gates do not suppress an occlusion that independently satisfies \eqref{eq:occlusion}.

The two brightness thresholds cannot be tuned independently. If $T_{\mathrm{sup}}>0$ and $\theta_{\mathrm{rapid}}\le\theta_{\mathrm{chg}}$, every frame satisfying $\delta_t>\theta_{\mathrm{chg}}$ also starts the rapid-brightness window before the scene decision. The primary scene path is therefore unreachable on that frame. We impose the consistency rule
\begin{equation}
T_{\mathrm{sup}}=0\quad\lor\quad
\theta_{\mathrm{rapid}}>\theta_{\mathrm{chg}}.
\label{eq:consistency}
\end{equation}
Persistence $K>1$ confirms a multi-frame transition, not a single step followed by a stable new view.

\subsection{Episode State, Complexity, and Invariants}
Two binary episode states select per-frame or once-per-episode emission; a clean decision ends the episode and updates all clean references, and scene decisions precede primary occlusion decisions. Mean extraction costs $\mathcal{O}(HW/4)$, gradient extraction $\mathcal{O}(HW/16)$, and the fixed grid $\mathcal{O}(32^2)$; storage is $\mathcal{O}(N)$ scalars plus $\mathcal{O}(N\cdot32^2)$ bytes. Three invariants make the detector auditable in a larger pipeline: \emph{reference safety} (no alarming frame is ever inserted into the scalar or grid reference, so a sustained occlusion cannot be absorbed by the baseline), \emph{episode non-duplication} (later positive frames belong to the active episode until a clean recovery frame resets it), and \emph{scene-path reachability} (the consistency rule of Eq.~\eqref{eq:consistency}).

\section{Controlled Evaluation}
Table~\ref{tab:eval_map} maps every evaluation protocol in this paper to its data source, unit of analysis, and the question it answers; the same method row can therefore appear under different scopes without those scopes being comparable.

\begin{table}[t]
\caption{Evaluation map. Primary in-scope evidence is separated from boundary and transfer diagnostics; metrics from different rows are not directly comparable.}
\label{tab:eval_map}
\centering
\scriptsize
\setlength{\tabcolsep}{2.6pt}
\begin{tabular}{p{0.27\columnwidth}p{0.24\columnwidth}p{0.14\columnwidth}p{0.25\columnwidth}}
\toprule
Protocol & Data & Units & Question\\
\midrule
Controlled stress (Table~\ref{tab:controlled}) & 8 background videos, fixed synthetic magnitudes & 400 seq. & primary in-scope comparison\\
Sensitivity grid (suppl.) & same & 400 seq. & parameter sensitivity\\
Extended stress, probes (suppl.) & synthetic gate stressors & 30--60 seq.\ each & per-gate behavior\\
Xiph audits & public videos & frames / seq. & reproducibility, latency\\
Magnitude sweep (Fig.~\ref{fig:magnitude_detectability}) & public videos, swept magnitudes & 250 seq. & decoupled operating characteristics\\
IoT cross-check (suppl.) & public stills, 2 cameras & images & untuned cross-camera transfer\\
UHCTD diagnostic (suppl.) & public video, 2 cameras & 4.5M frames & real-video boundary, in-scope stratification\\
Long-negative audit (suppl.) & VIRAT + CDnet2014 negatives & 9.09 cam.-h & verified-negative false alarms\\
\bottomrule
\end{tabular}
\end{table}

\subsection{Protocol and Baselines}
Ten random 48-frame clips are drawn from each of eight background videos (every third source frame, resized to $160\times90$), and five deterministic conditions are generated per clip, yielding 400 labeled sequences: clean, texture-collapsing cover, photometric scene replacement, geometry-preserving rotation/translation, and a global illumination step. Cover and photometric replacement are the declared in-scope positives; clean and illumination-step sequences are negatives; the geometric shift is an out-of-scope diagnostic that tests the absence of registration evidence. Perturbations begin at frame 16, after a 12-frame reference window; the seed is fixed to 20260701.

We compare six methods on identical sequences. The five baselines are transparent, deployable without training, and representative of the mean, gradient, edge, frame-difference, and histogram cues used in lightweight tamper detection: \emph{mean jump} (any post-onset $|\mu_t-\mu_{t-1}|>80$), \emph{gradient collapse} ($g_t<0.45g^\star$ after warm-up), an \emph{edge-reference proxy} (a literature-motivated Sobel edge-disappearance baseline without object masks, not claimed as an exact reproduction of \cite{lee2015unified}), \emph{frame difference} (mean absolute difference from the clean reference frame above 35 levels), and \emph{histogram difference} (Bhattacharyya distance from the clean reference histogram above 0.35). Complete published detectors such as Saglam and Temizel's adaptive scheme \cite{saglam2009adaptive} depend on pixelwise background maintenance outside the bounded-state family studied here and are represented by their constituent cues. The \emph{full state machine} runs Eqs.~\eqref{eq:scene}--\eqref{eq:coarse} with $N=12$, $\theta_{\mathrm{occ}}=0.45$, $\theta_{\mathrm{chg}}=80$, $\Delta_{\max}=45$, $[\mu_{\min},\mu_{\max}]=[60,120]$, $K=1$, $\rho_{\mathrm{grad}}=0.2$, and rapid suppression disabled to satisfy \eqref{eq:consistency} without masking scene alarms.

Performance is evaluated once per sequence. Primary metrics are computed only on the declared in-scope conditions: cover and photometric replacement as positives, and clean and illumination step as negatives. The geometric-shift condition is reported as an alarm-rate diagnostic rather than as a positive class in the primary metric.

\begin{table*}[t]
\caption{Controlled stress-test results. Of the 400 generated sequences (five conditions, 80 sequences each), the 320 primary-scope sequences (cover and photometric replacement as positives; clean and illumination step as negatives) define the aggregate metrics, with geometric shift retained only as an out-of-scope boundary diagnostic. Condition columns report alarm rates over 80 sequences per condition.}
\label{tab:controlled}
\centering
\setlength{\tabcolsep}{3.4pt}
\begin{tabular}{lcccccccccc}
\toprule
Method & Prec. & Recall & F1 & BAcc & FPR & Clean & Cover & Photo. & Illum. & Geom.\\
\midrule
Mean jump & 0.898 & 0.331 & 0.484 & 0.647 & 0.038 & 0.038 & 0.000 & 0.663 & 0.038 & 0.038\\
Gradient collapse & 0.927 & 0.637 & 0.756 & 0.794 & 0.050 & 0.050 & 0.963 & 0.313 & 0.050 & 0.050\\
Edge-reference proxy & 0.906 & 0.662 & 0.765 & 0.797 & 0.069 & 0.050 & 1.000 & 0.325 & 0.088 & 0.100\\
Frame difference & 0.602 & 0.994 & 0.750 & 0.669 & 0.656 & 0.313 & 0.988 & 1.000 & 1.000 & 0.963\\
Histogram difference & 0.635 & 1.000 & 0.777 & 0.713 & 0.575 & 0.150 & 1.000 & 1.000 & 1.000 & 0.213\\
Full state machine & 0.912 & 0.713 & \textbf{0.800} & \textbf{0.822} & 0.069 & 0.038 & 0.938 & 0.488 & 0.100 & 0.038\\
\bottomrule
\end{tabular}
\end{table*}

\subsection{Statistical Analysis}
All comparisons are paired by construction. On the 320 primary-scope sequences, the strongest baseline by F1 is histogram difference; the full detector alone is correct on 81 sequences and histogram difference alone on 46, giving exact McNemar $p=0.0024$, while the paired-bootstrap F1 difference is 0.023 with 95\% CI $[-0.041,0.085]$ (10,000 resamples). We therefore interpret the primary result as significantly improved paired correctness and false-alarm control on the declared scope, while the aggregate F1 advantage is not statistically resolved. The geometric-shift condition is excluded from the primary metric and retained as a boundary diagnostic: the full detector alarms on 3 of 80 such sequences (0.038) versus 77 of 80 for frame difference (0.963), showing why a moved-camera claim requires a registration cue.

\subsection{Interpretation}
Mean jump is selective but misses cover; gradient collapse detects most covers but reacts to nuisance conditions; the edge proxy is more balanced but remains a proxy without foreground masks; the frame- and histogram-difference detectors buy high recall with excessive negative-sequence alarms (31.3\% clean-sequence alarms for frame difference); the complete state machine gives the best primary-scope F1 and balanced accuracy at substantially lower FPR.

A basic ablation and threshold-sensitivity grid over the same 400 sequences (full table in the supplementary material) shows the expected monotone behavior: disabling the auxiliary luminance route trades recall for the lowest FPR (0.025); lowering the scene threshold to 60 doubles nuisance false alarms (FPR 0.200) while raising it to 100 costs photometric-transition recall; relaxing the occlusion ratio to 0.35 is the only single-parameter change that improves F1 (0.674 vs.\ 0.636 at the all-condition scope, which also counts geometric shift as a positive; see the supplementary material). The clean-only reference update and coarse-grid rejection variants produced no measurable change on this five-condition grid, so those design choices are instead isolated with the targeted stress probes summarized below.

\subsection{Extended Stress, Latency, and FPR-Constrained Operation}
An extended controlled stress audit with six additional generated conditions (sustained cover, repeated cover episodes, structured lighting, day--night transition, rapid brightness, close foreground) exercises the gates individually; the supplementary material reports the full component-triggered and targeted-probe tables. The component-level findings are: removing rapid-brightness suppression raises rapid-brightness false alarms from 0.217 to 0.767 and introduces day--night false alarms at 0.233; clean-only reference maintenance preserves a sustained-cover alarm that all-frame updating silently absorbs; the coarse-grid gate suppresses every structured-light false alarm that fires without it; and a lightweight ORB registration diagnostic detects all sampled geometric-shift clips but with 0.50 clean-clip FPR --- complementary, not deployable. Onset latency of detected positives is concentrated at zero frames (median and P90 zero, maximum 10, over 103 detections). A threshold sweep over the 360 extended-stress sequences (120 sustained- and repeated-cover positives, 240 nuisance negatives) reaches recall 0.925 at FPR $\le0.05$ using $\theta_{\mathrm{chg}}=150$ and $\theta_{\mathrm{occ}}=0.50$, with achieved FPR 0.025; because the sweep reuses generated data, we treat it as an FPR-constrained sensitivity diagnostic rather than a held-out operating point, and a field deployment should select thresholds on calibration cameras.

\subsection{Public-Data Reproducibility Audit}
Running the controlled protocol on eight small public Xiph.org test videos \cite{xiphDerf} produced 100 labeled synthetic stress sequences, on which the full state machine obtained precision 1.000, recall 0.500, F1 0.667, and FPR 0.000 (exact 95\% upper bound 0.071 on 0/50), while the high-recall frame-difference baseline again had higher F1 (0.730) at a much higher FPR (0.500) --- the same qualitative pattern as the local benchmark. Treating the unmodified public videos as clean negatives (808 frames), the detector emitted no alarms at 0.15 ms median per-frame simulator latency; a bounded 720p smoke test with unchanged thresholds also preserved zero false positives (details and intervals in the supplementary material). These are public-data sanity checks only and do not replace camera-disjoint field annotations or false alarms per camera-day.

\subsection{Magnitude-Swept Public Detectability Audit}
\label{sec:magnitude_sweep}
The controlled protocol fixes perturbation magnitudes near the detector's configured operating point, which risks overstating headline metrics. To remove this generator--detector coupling, we repeated the protocol on the public Xiph backgrounds with magnitudes sampled to \emph{cross} the configured thresholds: cover mean shifts in $[5,80]$ with opacity in $[0.6,1.0]$ (versus the accepted $(20,45]$), scene shifts in $[20,160]$ (versus $\theta_{\mathrm{chg}}=80$), and illumination steps in $[20,110]$. Each of the resulting 250 sequences records a continuous score per method, and the state machine is additionally swept along a monotone fourteen-point threshold path, yielding operating characteristics for all six methods (full curves in the supplementary material). At the false-alarm budget $\mathrm{FPR}\le0.05$ used throughout the paper, the full state machine attains the highest normalized partial AUC, 0.494, ahead of gradient collapse (0.422), histogram difference (0.394), the edge-reference proxy (0.380), mean jump (0.338), and frame difference (0.272). Because magnitudes now cross the declared envelope, absolute recall is necessarily lower than in Table~\ref{tab:controlled}; the informative quantity is the ranking under a fixed false-alarm budget.

\begin{figure*}[t]
\centering
\includegraphics[width=0.88\textwidth]{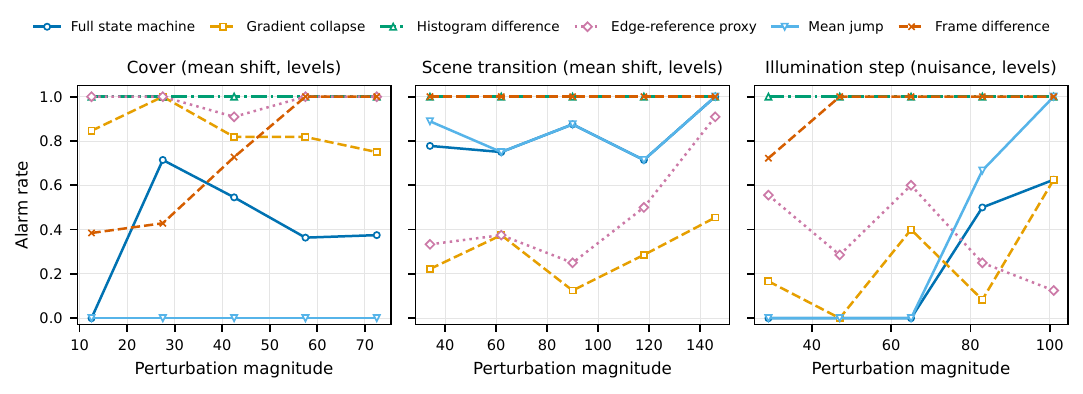}
\caption{Detection rate versus perturbation magnitude on the magnitude-swept public audit (five bins per condition, default thresholds). Cover detection by the full state machine peaks inside the accepted mean-shift interval $(20,45]$ and declines beyond $\Delta_{\max}$, which visualizes the declared operating envelope; the illumination panel shows the nuisance-rejection margin below $\theta_{\mathrm{chg}}$.}
\label{fig:magnitude_detectability}
\end{figure*}

Figure~\ref{fig:magnitude_detectability} reports detection rate as a function of perturbation magnitude: cover detection peaks inside the accepted $(20,45]$ mean-shift interval and declines beyond $\Delta_{\max}$ (making the declared envelope directly visible), and the state machine stays silent on illumination steps until the magnitude approaches $\theta_{\mathrm{chg}}$, whereas the high-recall frame- and histogram-difference baselines alarm on essentially every nuisance step.

\subsection{Independent Public Cross-Dataset Check}
\label{sec:iot_crosscheck}
The audits above still use synthetic perturbations. To test transfer to data we did not collect or design, we ran a single-shot check on the Bremen IoT image set, an independently collected, publicly released outdoor collection from two camera locations \cite{attarhaIoT2026}. These are periodic still captures rather than a continuous stream, so persistence, latency, and mode gating do not apply; we evaluate only the static occlusion predicate (Eq.~\eqref{eq:occlusion}) against a fixed reference built from a small sample of normal images, with no threshold retuned from the controlled-evaluation setting.

The set labels four classes per location: normal, blurred, rotated, and obstructed-lens (named \emph{noimage} in the dataset release). We report two protocols: within-camera, where reference and test images come from the same location, and camera-disjoint, where the reference from one location is applied unmodified to the other. Across all four protocol instances (per-protocol and per-method tables in the supplementary material), the occlusion predicate held false positives on held-out normal images to 0.000--0.009, at the cost of recall of 0.065--0.199 concentrated on the obstructed-lens class (per-category alarm rate 0.195--0.386), a weak response to blur (0.00--0.28), and essentially no response to rotation ($\le$0.006) --- exactly the declared operating envelope of Section~III. The three reference-based baselines transferred far worse under the camera-disjoint protocol, with false-positive rates of 0.15--1.00; the same clean-only reference logic that limits recall here is what kept the false-alarm side stable without per-camera recalibration. 
Representative frames behind these numbers appear in the supplementary material: the flagged full-lens obstruction is a near-uniform frame with collapsed gradient, a clean match to Eq.~\eqref{eq:occlusion}; a missed defocus case retains enough residual luminance structure to fall outside the required mean-shift interval; and a missed rotation case is nearly indistinguishable from the normal reference in global mean and sampled gradient, grounding the declared out-of-scope rotation case in a real frame.

\subsection{Public UHCTD Video Diagnostic}
\label{sec:uhctd_diagnostic}
UHCTD is a public multi-day camera-tampering video dataset with normal, covered, defocused, and moved-camera conditions and published deep-network predictions \cite{mantini2019uhctd}, used here as an external diagnostic. Frame class 0 is treated as negative and classes 1--3 as positive; this binary mapping is deliberately broader than our operating envelope (camera movement is out of scope and defocus only weakly covered), and no thresholds were tuned on UHCTD.

The full protocol tables appear in the supplementary material. The official released Camera~A predictions over Days~3--6 (AlexNet experiments) reach recall~0.86--0.88 but at normal-frame FPR~0.28--0.29; the release provides hard class labels rather than scores, so these baselines cannot be re-thresholded to the FPR~$\le0.05$ budget used elsewhere in this paper. A full decoded-video pass (stride~1) of both cameras shows the full state machine is not completely silent on this domain --- recall~0.028 at FPR~0.049 on Camera~A (1,032,624 frames) and recall~0.021 at FPR~0.039 on Camera~B (3,454,272 frames) --- far below the official CNN baselines. The transparent baselines buy recall only with false alarms: gradient collapse reaches recall~0.36--0.49 at FPR~0.44--0.53, and histogram difference recall~0.67--0.90 at FPR~0.57--0.78. A frozen ResNet18 feature-distance baseline on a 1,200-frame sample reaches F1~0.681 at its best-F1 threshold and recall~0.392 when constrained to FPR~$\le0.05$. Stratifying all 576 labeled events against the operating envelope explains this: only 22 fall in scope, and those reach recall~0.409 (0.667 for covers) versus recall~0.016 for the 382 out-of-scope events (exact intervals in the supplementary material) --- detection concentrates where the envelope predicts.

To test whether a calibrated learned model can recover recall, we fine-tuned the last layer of ResNet18 as a binary tamper classifier on Camera~A Days~3--4: the same-camera held-out split (Days~5--6) reaches recall~0.890 at FPR~0.049 (F1~0.889 at its best-F1 threshold), while the camera-disjoint split (Camera~B) collapses --- its best-F1 point already saturates FPR near~1.0, and no threshold recovers nonzero recall at FPR~$\le0.05$. Calibration therefore transfers only within the training camera, and UHCTD serves overall as an external operating-boundary diagnostic: the full state machine detects a small fraction of UHCTD events (primarily severe covers satisfying the mean-shift interval) at moderate FPR, and calibration or learned baselines are needed before extending the claim to moved-camera and defocus classes.

We also ran a bounded camera-disjoint threshold-transfer diagnostic: FPR~$\le0.05$ thresholds selected on one UHCTD camera do not transfer reliably to the other in either direction (protocol and table in the supplementary material), supporting the need for a larger camera-disjoint calibration protocol before any broad UHCTD or field claim.

\section{Discussion and Limitations}
The main operational advantage is a small, auditable decision system --- tampered frames cannot poison the reference, nuisance gates have explicit semantics, and episode state prevents repeated notifications --- supported by the primary controlled results on its matched scope, with the geometric diagnostic and external checks defining where registration, calibration, or learned references are required. The public-data evidence should be read across four sampling regimes rather than as one pooled rate: the Xiph audit checks reproducibility with synthetic stress and clean playback; the independent image check shows near-zero normal-image false alarms under camera-disjoint transfer (0.000--0.009), supporting the clean-reference transfer story for still images; and the UHCTD full decoded pass (4.5M frames) locates the boundary where fixed predicates detect only a small fraction of events (recall 0.021--0.028 at FPR 0.039--0.049) while a fine-tuned ResNet18 (F1~0.889 same-camera) collapses to near-zero recall at the same budget once transferred across cameras, motivating calibration or learned baselines before broader claims. The fourth check --- a public long-negative audit spanning VIRAT \cite{oh2011virat} and four CDnet2014 \cite{wang2014cdnet} categories (345 recordings, 9.09 camera-hours; table in the supplementary material) --- records zero false alarms for the full state machine, bounding its rate below 0.33 false alarms per camera-hour (one-sided 95\%; recording-length caveat in the supplementary material). The supplementary operational candidate pools stay out of the main false-positive evidence --- they lack long-negative protocols and several rows use surrogate labels; their diagnostic value is exposing an enriched-pool false-alarm rate and close-range cover false negatives, motivating camera-disjoint calibration and predicate-level remedies.

Several limitations remain. First, a held-out, independently annotated real-camera test set with physically executed tamper events, onset/offset labels, and day/night metadata is still required; no public dataset currently provides physically executed tamper events in continuous video (UHCTD injects tampering synthetically; the Bremen set provides physical tampers only as still images), so a scripted multi-camera field study remains future work. Second, pure geometric displacement needs a complementary registration cue, and subtle partial covers or mean-preserving defocus violate Eq.~\eqref{eq:occlusion}. Third, fixed thresholds remain camera dependent; percentile calibration should be evaluated against manual settings, though the false-alarm side of this concern did not materialize under untuned transfer in Section~\ref{sec:iot_crosscheck}. Finally, the contribution is hardware-agnostic: latency and throughput were measured on commodity CPU hardware, and a dedicated embedded target is not a prerequisite for the algorithmic claims.

As a supplementary systems-level data point, the standalone C++ implementation processes 10,000 synthetic frames at $160{\times}90$ in 0.32~s (31,400~fps) and 1,000 frames at $1280{\times}720$ in 0.79~s (1,270~fps) on a commodity x86-64 CPU; because the state machine subsumes mean, gradient, and grid computations, this upper-bounds the per-frame cost of every baseline.

\section{Reproducibility, Data, and Deployment Boundaries}
All evaluations use public data (Xiph, Bremen IoT, UHCTD, VIRAT, CDnet2014) or a fully specified protocol, except the primary benchmark's unreleased background video (public-data equivalent in Section~V-D); the supplementary material gives the exact predicates, configuration, and statistical procedures, and the release artifact will include scripts reproducing every reported public-data table and figure.

Auditability requires preserving diagnostic metadata at the alarm frame, not just the final decision. A minimal per-frame record is
\begin{equation}
m_t=(\mu_t,g_t,\mu_t^\star,g_t^\star,\delta_t,L_t,s_t,\tau_t),
\end{equation}
where $L_t$ is the coarse-grid rejection flag, $s_t$ the suppression/mode state, and $\tau_t$ the episode state: the same gradient collapse is triaged as a cover candidate when $L_t=0$ and as a structured-lighting rejection when $L_t=1$; archive this record for false-alarm review. A rigorous field evaluation should publish camera-disjoint splits, attack types with onset/offset, day/night state, tuning provenance, per-type recall, onset latency, false alarms per camera-day, and paired confidence intervals; long negative recordings qualify only under the definition of Section~\ref{sec:ground-truth-definition}.

Because the domain is surveillance video, released examples are screened for privacy and permission; the detector is a camera-health signal for maintenance --- not evidence about people, behavior, or intent --- and deployment should retain human review and site-specific calibration given the deliberately excluded tamper modes.

\section{Conclusion}
We presented an implementation-faithful, configuration-aware camera-tamper state machine for texture-collapsing occlusion and abrupt photometric scene transition, in which clean-only references, coarse structural rejection, transition suppression, and episode control yield deterministic behavior at low arithmetic cost. A controlled paired experiment shows improved primary-scope F1 and balanced accuracy with significantly better paired correctness than the strongest transparent baseline (the paired F1 interval still overlaps zero), the magnitude-swept public audit adds the highest partial AUC under a 5\% false-alarm budget, an in-scope stratification of 4.5M real frames confirms envelope-consistent detection, and a verified-negative public audit records zero false alarms in 9.09 camera-hours. The contribution is deliberately an auditable sensor-health subsystem for two low-cost signatures rather than a general classifier, with moved-camera detection delegated to a registration cue and field calibration still required before broad deployment claims.

\bibliographystyle{IEEEtran}
\bibliography{references}

\end{document}


\maketitle

\section{Implementation Traceability and Exact Predicates}
The primary input is an 8-bit single-channel frame. Mean luminance samples every fourth column on every row, whereas the gradient samples every fourth row and column and uses the immediate right and lower neighbors. Consequently, their leading costs are $HW/4$ samples and approximately $2HW/16$ absolute differences, respectively.

The current single-frame pathway exposes an auxiliary luminance/gradient interface with
\begin{equation}
\ell_t=1,\qquad \mu_t^{\mathrm{aux}}=\mu_t,\qquad
g_t^{\mathrm{aux\,enabled}}=0.
\end{equation}
Thus the auxiliary valid-ratio occlusion test cannot fire, the auxiliary-gradient test is disabled, and the only active auxiliary contribution is an additional mean-change comparison against the last accepted clean frame and the sliding reference. The coarse $32\times32$ luminance grid is likewise not a positive detector. It is evaluated only after the scalar occlusion predicate has fired and can only reject that candidate as a structured lighting change.

Let $\mu_t^\star$ and $g_t^\star$ be the clean-window means, $\delta_t=|\mu_t-\mu_{t-1}|$, and $\epsilon=10^{-6}$. The primary scene predicate before persistence and suppression is
\begin{align}
\mathcal{S}_t={}&[\delta_t>\theta_{\mathrm{chg}}]
\land[\neg o_t\lor o_{t-1}]\nonumber\\
&\land\left[\rho_{\mathrm{grad}}\le0\ \lor\
\frac{g_{t-1}-g_t}{\max(g_t^\star,\epsilon)}
\ge\rho_{\mathrm{grad}}\right],
\end{align}
where $o_j=[g_j<\theta_{\mathrm{occ}}g_t^\star]$. A counter increments only on consecutive $\mathcal{S}_t=1$ frames and resets otherwise.

The primary occlusion predicate is
\begin{align}
\mathcal{O}_t={}&[g_t<\theta_{\mathrm{occ}}g_t^\star]
\land[|\mu_t-\mu_t^\star|>20]\nonumber\\
&\land[|\mu_t-\mu_t^\star|\le\Delta_{\max}]
\land[\mu_{\min}\le\mu_t\le\mu_{\max}]
\land[\neg L_t],
\label{eq:supp-occlusion}
\end{align}
where $L_t$ is the coarse-grid structured-lighting predicate described in the main paper.

For the active auxiliary luminance route, let $\mu_{t^-}$ be the last accepted clean-frame mean. Then
\begin{align}
\mathcal{A}_t={}&
[|\mu_t-\mu_{t^-}|>\theta_{\mathrm{chg}}\ \lor\
|\mu_t-\mu_t^\star|>\theta_{\mathrm{chg}}]\nonumber\\
&\land\left[
\frac{|\mu_t-\mu_{t^-}|}{\max(\mu_{t^-},1)}>\rho_{\mathrm{lumin}}
\ \lor\
\frac{|\mu_t-\mu_t^\star|}{\max(\mu_t^\star,1)}>\rho_{\mathrm{lumin}}
\right].
\end{align}
This route is evaluated only outside mode-change and rapid-brightness suppression windows. It may reinforce a scene decision but does not overwrite an already selected occlusion.

\section{Suppression Consistency Proof}
Suppose $T_{\mathrm{sup}}>0$ and $\theta_{\mathrm{rapid}}\le\theta_{\mathrm{chg}}$. For any frame satisfying the first conjunct of the primary scene predicate,
\begin{equation}
\delta_t>\theta_{\mathrm{chg}}\ge\theta_{\mathrm{rapid}}.
\end{equation}
The rapid-change timestamp is updated before the scene predicate is evaluated. At that same timestamp,
\begin{equation}
0\le \tau_t-\tau_{\mathrm{rapid}}=0<T_{\mathrm{sup}},
\end{equation}
so scene output is suppressed. Therefore the primary scene path cannot emit on its triggering frame. The sufficient configuration rule is $T_{\mathrm{sup}}=0$ or $\theta_{\mathrm{rapid}}>\theta_{\mathrm{chg}}$. This rule does not guarantee scene recall; it only removes the deterministic masking contradiction.

\section{Reference and Episode Update Order}
\begin{algorithm}[t]
\caption{Streaming Decision Order}
\begin{algorithmic}[1]
\STATE Validate single-channel frame geometry.
\STATE Compute mean, gradient, and coarse sample.
\STATE Update rapid-brightness timestamp and suppression state.
\IF{clean reference is not initialized}
\STATE Emit but do not store a boot-occluded frame.
\STATE Otherwise insert the frame into warm-up buffers.
\STATE \textbf{return}.
\ENDIF
\STATE Evaluate the primary scene predicate and persistence.
\IF{no primary scene decision}
\STATE Evaluate scalar occlusion and coarse-grid rejection.
\ENDIF
\STATE Evaluate the active auxiliary luminance scene route.
\STATE Emit according to per-frame or one-per-episode policy.
\IF{no tamper decision}
\STATE Update scalar, auxiliary-luminance, and grid references.
\ENDIF
\STATE Update previous-frame mean and gradient on every frame.
\STATE Update previous auxiliary mean only on an accepted clean frame.
\end{algorithmic}
\end{algorithm}

\section{Controlled Benchmark Details}
Table~\ref{tab:config} records the configuration used for the reported controlled experiment. Rapid suppression was disabled because the experiment measures the photometric scene path directly; deployments that enable it must obey the consistency condition in the main paper.

\begin{table}[t]
\caption{Controlled-Evaluation Configuration}
\label{tab:config}
\centering
\resizebox{\columnwidth}{!}{%
\begin{tabular}{lc}
\toprule
Parameter & Value\\
\midrule
Clean window $N$ & 12\\
Occlusion gradient ratio & 0.45\\
Scene mean jump & 80\\
Minimum/maximum mean shift & 20/45\\
Accepted occlusion mean interval & [60, 120]\\
Scene persistence & 1 frame\\
Required gradient change ratio & 0.2\\
Auxiliary relative mean ratio $\rho_{\mathrm{lumin}}$ & 0.8\\
Rapid-brightness threshold $\theta_{\mathrm{rapid}}$ (intensity levels) & 1000\\
Suppression window $T_{\mathrm{sup}}$ (ms) & 0 (disabled)\\
Boot dark / flat thresholds & 15 / 0 (flat check disabled)\\
Clean-only reference update & enabled\\
Coarse-grid rejection & enabled\\
Auxiliary luminance route & enabled\\
Input size / clip length & $160\times90$ / 48 frames\\
Random seed & 20260701\\
\bottomrule
\end{tabular}
}
\end{table}

The cover perturbation blends the post-onset frame with a near-uniform surface whose mean differs from the warm-up reference by 24--40 levels and adds Gaussian noise with standard deviation 1.2. The photometric scene perturbation replaces the post-onset content with frames from a different source video and shifts mean luminance by 95 levels. The geometric stress test applies an 18-degree rotation, a horizontal translation of 12\% of width, and a vertical translation of 8\% of height with reflected borders. The nuisance illumination condition adds or subtracts 70 levels without changing geometry.

\section{Declared Operating Envelope}
Table~\ref{tab:failure_modes} tabulates the operating envelope declared in the
main paper, condition by condition, with the mechanistic reason for each
expected behavior. It reports method semantics, not new experimental
evidence.

\begin{table*}[t]
\caption{Declared operating envelope and expected failure modes.}
\label{tab:failure_modes}
\centering
\scriptsize
\setlength{\tabcolsep}{4.0pt}
\begin{tabular}{p{0.23\textwidth}p{0.24\textwidth}p{0.43\textwidth}}
\toprule
Condition & Expected detector behavior & Reason\\
\midrule
Texture-collapsing cover with bounded mean shift & In scope & Gradient collapse and bounded luminance change satisfy the occlusion predicate.\\
Abrupt photometric scene transition & In scope when not suppressed by mode or rapid-brightness gates & Inter-frame mean jump and gradient-change checks drive the scene path.\\
Pure camera rotation or translation & Out of scope & Global mean and sampled local-gradient statistics can remain close to the clean reference without registration evidence.\\
Mean-preserving defocus & Out of scope or weak response & Defocus may reduce texture without satisfying the required mean-shift interval.\\
Subtle local partial cover & Out of scope or weak response & Sampling and global aggregate statistics may dilute small localized changes.\\
Global illumination step & Intended nuisance rejection & Rapid-brightness and structural-lighting logic are designed to avoid turning lighting transients into tamper episodes.\\
Foreground object close to lens & Ambiguous, requires field validation & The current predicates do not include semantic foreground reasoning or object masks.\\
Unreadable or failed clip & Excluded from accuracy metrics & Execution failure is an input-quality or processing-failure case, not a true negative or false negative without annotation.\\
\bottomrule
\end{tabular}
\end{table*}

\section{Magnitude-Sweep Operating Characteristics}
\begin{figure}[t]
\centering
\includegraphics[width=\columnwidth]{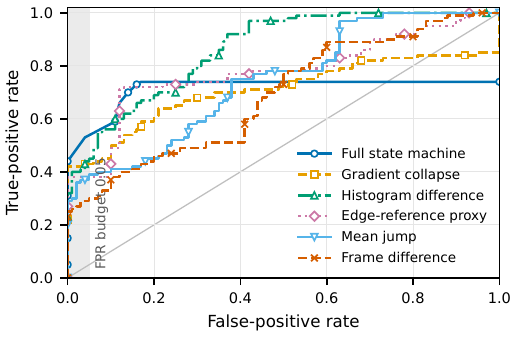}
\caption{Operating characteristics on the magnitude-swept public audit (250 sequences, primary scope). The shaded band marks the $\mathrm{FPR}\le0.05$ false-alarm budget. Baselines are swept over their native scalar statistics; the full state machine is swept along a monotone fourteen-point threshold path, so its curve ends at the recall ceiling of its most sensitive configured setting.}
\label{fig:magnitude_roc}
\end{figure}

Figure~\ref{fig:magnitude_roc} shows the full operating-characteristic curves
for the magnitude-swept public audit summarized in the main paper.

\section{Ablation and Threshold-Sensitivity Grid}
Table~\ref{tab:ablation_sensitivity} reports the basic ablation and threshold-sensitivity grid over the full 400-sequence controlled set summarized in the main paper. The clean-only reference update and coarse-grid rejection variants produced no measurable change on this five-condition grid; those design choices are isolated with the targeted stress probes of the next section instead.

\begin{table}[t]
\caption{Controlled ablation and threshold-sensitivity summaries on the same 400 sequences. Condition columns report alarm rate. Aggregate columns use the all-condition scope (cover, photometric replacement, and geometric shift as positives; clean and illumination step as negatives), which is why the Full state machine row differs from the primary-scope metrics of Table~II in the main paper.}
\label{tab:ablation_sensitivity}
\centering
\scriptsize
\setlength{\tabcolsep}{2.6pt}
\begin{tabular}{lcccccccc}
\toprule
Variant & Prec. & Recall & F1 & BAcc & FPR & Cover & Photo. & Illum.\\
\midrule
Full state machine & 0.914 & 0.488 & 0.636 & 0.709 & 0.069 & 0.938 & 0.488 & 0.100\\
No auxiliary luminance & 0.963 & 0.429 & 0.594 & 0.702 & 0.025 & 0.938 & 0.350 & 0.050\\
Persistence $K=2$ & 0.911 & 0.471 & 0.621 & 0.701 & 0.069 & 0.938 & 0.438 & 0.100\\
Occlusion ratio 0.35 & 0.927 & 0.529 & 0.674 & 0.733 & 0.063 & 0.938 & 0.613 & 0.088\\
Scene threshold 60 & 0.797 & 0.525 & 0.633 & 0.663 & 0.200 & 0.938 & 0.563 & 0.350\\
Scene threshold 100 & 0.910 & 0.421 & 0.575 & 0.679 & 0.063 & 0.938 & 0.288 & 0.088\\
Max mean shift 35 & 0.929 & 0.488 & 0.639 & 0.716 & 0.056 & 0.938 & 0.488 & 0.075\\
Persistence $K=3$ & 0.911 & 0.471 & 0.621 & 0.701 & 0.069 & 0.938 & 0.438 & 0.100\\
\bottomrule
\end{tabular}
\end{table}

\section{Extended Stress and Targeted Component Diagnostics}
The extended controlled stress audit adds six generated conditions to the base protocol: sustained cover, repeated cover episodes, structured lighting, day--night transition, rapid brightness, and close foreground. Table~\ref{tab:extended_ablation} reports the component-triggered behavior that changed under this audit. Removing rapid-brightness suppression raises rapid-brightness false alarms from 0.217 to 0.767 and introduces day--night false alarms at 0.233.

\begin{table}[t]
\caption{Component-triggered extended stress audit. Entries show alarmed sequences over the condition denominator and the corresponding sequence-level alarm rate.}
\label{tab:extended_ablation}
\centering
\scriptsize
\setlength{\tabcolsep}{2.8pt}
\begin{tabular}{lccc}
\toprule
Variant & Long cover & Day--night & Rapid bright.\\
\midrule
Full state machine & 58/60 (0.967) & 0/60 (0.000) & 13/60 (0.217)\\
No auxiliary luminance & 58/60 (0.967) & 0/60 (0.000) & 0/60 (0.000)\\
No rapid suppression & 58/60 (0.967) & 14/60 (0.233) & 46/60 (0.767)\\
\bottomrule
\end{tabular}
\end{table}

Three targeted component probes and one geometric baseline diagnostic isolate the design choices that the five-condition grid cannot separate (Table~\ref{tab:targeted_diag}). In a sustained-cover contamination probe, clean-only reference maintenance preserves the alarm through the post-onset tail, whereas all-frame updating absorbs the cover and stops alarming. In a structured-light probe that satisfies the scalar occlusion predicate, the coarse-grid gate suppresses all alarms while the no-grid variant alarms on every sequence. In an operating-mode probe, the mode gate eliminates the day--night nuisance alarms that appear when the gate is removed. Finally, a lightweight ORB homography-failure diagnostic detects all sampled geometric-shift clips, confirming that a complementary registration cue directly targets the declared out-of-scope motion case, but its 0.50 FPR on clean clips shows that such a cue needs calibration; it demonstrates complementarity and is not part of the primary method comparison.

\begin{table}[t]
\caption{Targeted component and registration diagnostics. Rates are sequence-level alarm rates except for the ORB row, which reports binary detection metrics on six geometric-shift and six clean clips.}
\label{tab:targeted_diag}
\centering
\scriptsize
\setlength{\tabcolsep}{3.2pt}
\begin{tabular}{p{0.30\columnwidth}p{0.18\columnwidth}p{0.20\columnwidth}p{0.22\columnwidth}}
\toprule
Diagnostic & Full / proposed & Ablated / baseline & Interpretation\\
\midrule
Clean-only reference & 1.000 & 0.000 & Maintains sustained-cover alarm\\
Coarse-grid gate & 0.000 & 1.000 & Rejects structured-light false alarm\\
Mode-transition gate & 0.000 & 0.500 & Suppresses mode-switch nuisance\\
ORB registration & 1.000 recall & 0.500 FPR & Detects motion but is noisy\\
\bottomrule
\end{tabular}
\end{table}

The same audit gives onset-latency statistics for positive detections. The full state machine detects with median latency 0 frames, P90 latency 0 frames, maximum latency 10 frames, and mean latency 0.64 frames over 103 detected positives. Mean jump detects only three positives and alarms four frames after onset; frame difference has median/P90 latency 0 frames and maximum latency 9 frames. The FPR-constrained threshold sweep reported in the main paper runs on the 360 extended-stress sequences --- 120 positives (60 sustained-cover and 60 repeated-episode clips) and 240 nuisance negatives (60 each of structured lighting, day--night, rapid brightness, and close foreground) --- and achieves recall 0.925 at FPR 0.025 with the confusion counts 111 TP, 6 FP, 234 TN, 9 FN.

\section{High-Resolution Processing Smoke Test}
To check resolution sensitivity of the implementation path, the public-video backgrounds were resized to $1280\times720$ and re-run without changing thresholds. Because the source clips are low-resolution public test media, this is a high-resolution processing check rather than native HD field evidence. On 30 synthetic sequences, the full state machine obtained precision 1.000, recall 0.333, F1 0.500, BAcc 0.667, and FPR 0.000, with median per-frame simulator time 0.76 ms and 95th percentile 2.15 ms. The 720p FPR of 0/12 has a 95\% Clopper--Pearson upper bound of 0.265, and its recall of 6/18 has 95\% CI [0.133, 0.590]; it is therefore only a runtime and resolution-path smoke test.

\section{UHCTD Diagnostic Tables}
Table~\ref{tab:uhctd_diag} gives the UHCTD diagnostic summarized in the main
paper: official released Camera~A learned-model predictions, the learned
baselines, and the full decoded-video pass of the full state machine.
Table~\ref{tab:uhctd_full_baselines} gives the per-method transparent-baseline
rows for the same full decoded-video pass (stride~1), summarized as ranges in
the main paper.

\begin{table}[t]
\caption{UHCTD public-video diagnostic on Days 3--6. Official rows evaluate released Camera~A learned-model predictions on all aligned frames. Full-pass rows decode every frame (stride~1). Fine-tuned ResNet18 rows: same-camera trains and tests on Camera~A; cross-camera trains on Camera~A and tests on Camera~B. The cross-camera FPR$\le$0.05 row is degenerate (no threshold reaches the budget with nonzero recall, so recall and FPR both saturate at 0), reported to make the cross-camera failure explicit rather than omitted.}
\label{tab:uhctd_diag}
\centering
\scriptsize
\setlength{\tabcolsep}{3.2pt}
\begin{tabular}{llrrrr}
\toprule
Protocol & Method & Frames & Recall & FPR & F1\\
\midrule
Official full & Exp.~3 AlexNet & 1,032,000 & 0.863 & 0.283 & 0.640\\
Official full & Exp.~1 AlexNet & 1,032,192 & 0.876 & 0.293 & 0.639\\
Same-cam (best-F1) & Fine-tuned ResNet18 & 10,000 & 0.843 & 0.018 & 0.889\\
Same-cam (FPR$\le$0.05) & Fine-tuned ResNet18 & 10,000 & 0.890 & 0.049 & 0.875\\
Cross-cam (best-F1) & Fine-tuned ResNet18 & 20,000 & 1.000 & 1.000 & 0.387\\
Cross-cam (FPR$\le$0.05) & Fine-tuned ResNet18 & 20,000 & 0.000 & 0.000 & 0.000\\
Frozen sample & Frozen ResNet18 & 1,200 & 0.767 & 0.207 & 0.681\\
\midrule
Full pass (A) & Full state machine & 1,032,624 & 0.028 & 0.049 & 0.047\\
Full pass (B) & Full state machine & 3,454,272 & 0.021 & 0.039 & 0.038\\
\bottomrule
\end{tabular}
\end{table}

\begin{table}[t]
\caption{UHCTD full decoded-video pass, transparent baselines (Days 3--6).}
\label{tab:uhctd_full_baselines}
\centering
\scriptsize
\setlength{\tabcolsep}{3.2pt}
\begin{tabular}{llrrrr}
\toprule
Protocol & Method & Frames & Recall & FPR & F1\\
\midrule
Full pass (A) & Gradient collapse & 1,032,624 & 0.362 & 0.440 & 0.272\\
Full pass (B) & Gradient collapse & 3,454,272 & 0.488 & 0.529 & 0.319\\
Full pass (A) & Frame difference & 1,032,624 & 0.646 & 0.584 & 0.383\\
Full pass (B) & Frame difference & 3,454,272 & 0.601 & 0.526 & 0.380\\
Full pass (A) & Histogram difference & 1,032,624 & 0.897 & 0.784 & 0.426\\
Full pass (B) & Histogram difference & 3,454,272 & 0.668 & 0.569 & 0.398\\
\bottomrule
\end{tabular}
\end{table}

\section{UHCTD In-Scope Event Stratification}
Each of the 576 labeled UHCTD events (both cameras, Days 3--6) is classified against the declared operating envelope using a ground-truth-clean reference computed from the sliding mean of sampled statistics over the last 12 normal-labeled frames preceding the event: an event is envelope-in-scope if any frame within its first 300 frames satisfies the occlusion predicate (Eq.~\eqref{eq:supp-occlusion}: gradient ratio, mean-shift interval, and luma bounds all met) or the scene predicate (inter-frame mean jump exceeds the scene threshold); it is a boundary case if only the texture-collapse condition holds without the required mean-shift interval; otherwise it is out of scope. Table~\ref{tab:uhctd_inscope} joins this classification with the full state machine's alarm frames from the full decoded-video pass. In-scope covers are detected at recall~0.667, far above the aggregate rate, while out-of-scope events across all three tamper classes are detected at recall~0.011--0.022, confirming that the full state machine's low aggregate recall on UHCTD reflects selective, envelope-consistent blindness rather than a uniformly weak detector. Because the in-scope cells are small, we report exact 95\% Clopper--Pearson intervals: covered 6/9, $[0.30,0.93]$; in-scope overall 9/22, $[0.21,0.64]$; out-of-scope 6/382, $[0.006,0.034]$; the in-scope moved cell contains only three events and its rate should not be over-read. The in-scope and out-of-scope intervals do not overlap, so the qualitative conclusion survives the small-sample bound.

\begin{table}[t]
\caption{UHCTD in-scope stratified event recall, both cameras combined (576 labeled events).}
\label{tab:uhctd_inscope}
\centering
\scriptsize
\setlength{\tabcolsep}{4pt}
\begin{tabular}{llrrr}
\toprule
Scope & Class & Events & Detected & Recall\\
\midrule
In-scope & Covered & 9 & 6 & 0.667\\
In-scope & Defocused & 10 & 1 & 0.100\\
In-scope & Moved & 3 & 2 & 0.667\\
In-scope & All & 22 & 9 & 0.409\\
Boundary (texture only) & Defocused & 166 & 9 & 0.054\\
Boundary (texture only) & Moved & 6 & 0 & 0.000\\
Boundary (texture only) & All & 172 & 9 & 0.052\\
Out-of-scope & Covered & 183 & 2 & 0.011\\
Out-of-scope & Defocused & 16 & 0 & 0.000\\
Out-of-scope & Moved & 183 & 4 & 0.022\\
Out-of-scope & All & 382 & 6 & 0.016\\
\midrule
All & All & 576 & 24 & 0.042\\
\bottomrule
\end{tabular}
\end{table}

\section{UHCTD Camera-Disjoint Threshold Transfer}
For each transfer direction, 100 sampled frames per day were decoded from each camera over Days 3--6, giving 400 calibration frames and 400 held-out frames; each stream retained its own clean reference, but score thresholds were selected only on the calibration camera (FPR $\le0.05$ objective) and then frozen on the held-out camera. The sampled subset is diagnostic rather than exhaustive, but it directly tests whether a false-alarm-constrained threshold transfers across cameras. Table~\ref{tab:uhctd_transfer} shows that it does not transfer reliably: Camera A thresholds become over-conservative on Camera B, while Camera B thresholds can exceed the held-out false-alarm budget on Camera A.

\begin{table}[t]
\caption{UHCTD camera-disjoint threshold-transfer diagnostic. Thresholds are selected on the calibration camera with the FPR $\le0.05$ objective and then frozen on the held-out camera. Each direction uses 400 sampled calibration frames and 400 sampled held-out frames.}
\label{tab:uhctd_transfer}
\centering
\scriptsize
\setlength{\tabcolsep}{3.4pt}
\begin{tabular}{llrrr}
\toprule
Cal.$\to$Test & Method & Recall & FPR & F1\\
\midrule
A$\to$B & Mean jump & 0.000 & 0.009 & 0.000\\
A$\to$B & Frozen ResNet18 & 0.000 & 0.000 & 0.000\\
B$\to$A & Frame difference & 0.775 & 0.221 & 0.676\\
B$\to$A & Frozen ResNet18 & 1.000 & 0.786 & 0.522\\
\bottomrule
\end{tabular}
\end{table}

\section{Full Public Image Cross-Dataset Breakdown}
Table~\ref{tab:public_iot} gives the per-protocol summary of the independent public cross-dataset check, and Table~\ref{tab:public_iot_full_appendix} gives the full four-method breakdown summarized in the main paper.

\begin{table}[t]
\caption{Independent public cross-dataset check summary. $L_1$ and $L_2$ denote the two independent camera locations; cross-camera rows use an untuned reference from the source location. Baseline FPR range covers gradient-collapse, frame-difference, and histogram-difference baselines. \emph{Noimage} is the dataset's name for the obstructed-lens class.}
\label{tab:public_iot}
\centering
\scriptsize
\setlength{\tabcolsep}{3.0pt}
\begin{tabular}{lcccc}
\toprule
Protocol & Recall & FPR & Noimage & Baseline FPR\\
\midrule
Within $L_1$ & 0.065 & 0.000 & 0.195 & 0.340--0.349\\
Within $L_2$ & 0.199 & 0.000 & 0.243 & 0.323--0.680\\
$L_1\!\to\!L_2$ & 0.155 & 0.009 & 0.386 & 0.969--1.000\\
$L_2\!\to\!L_1$ & 0.140 & 0.000 & 0.278 & 0.151--1.000\\
\bottomrule
\end{tabular}
\end{table}

This breakdown is diagnostic rather than primary evidence: the static occlusion predicate keeps normal-image false alarms low under untuned camera transfer, while learned calibration, continuous-video persistence, and registration cues remain outside this still-image check.

\begin{figure}[t]
\centering
\begin{minipage}{0.48\columnwidth}
\centering
\includegraphics[width=\linewidth]{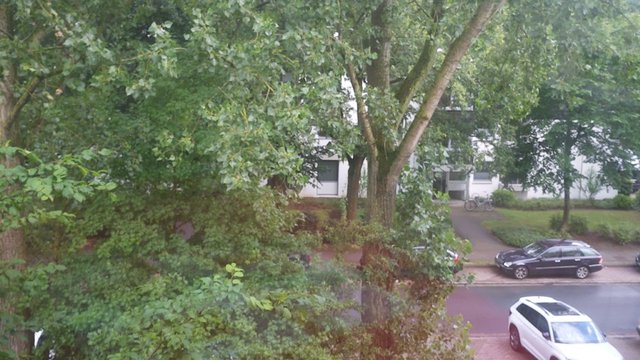}\\[2pt]
\footnotesize (a) Normal reference
\end{minipage}\hfill
\begin{minipage}{0.48\columnwidth}
\centering
\includegraphics[width=\linewidth]{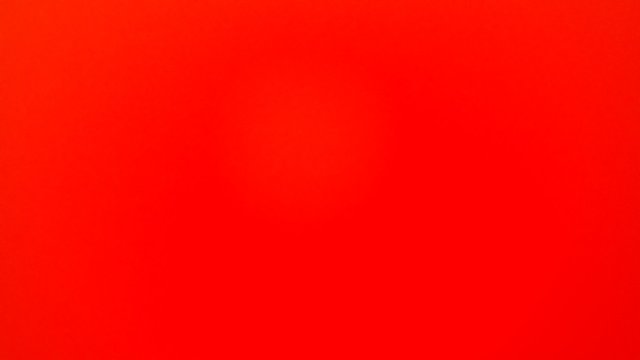}\\[2pt]
\footnotesize (b) Obstructed lens: detected
\end{minipage}

\vspace{4pt}

\begin{minipage}{0.48\columnwidth}
\centering
\includegraphics[width=\linewidth]{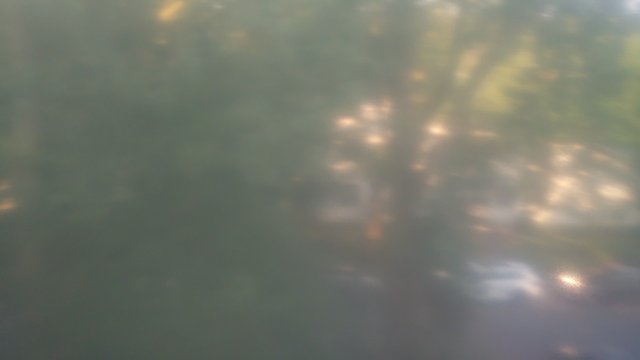}\\[2pt]
\footnotesize (c) Defocus: missed
\end{minipage}\hfill
\begin{minipage}{0.48\columnwidth}
\centering
\includegraphics[width=\linewidth]{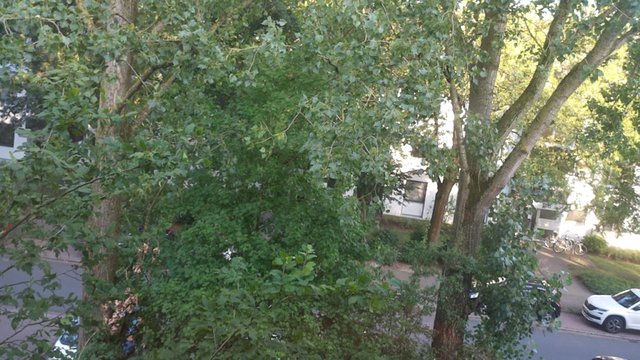}\\[2pt]
\footnotesize (d) Rotation: missed
\end{minipage}
\caption{Qualitative success and failure examples from the independent public cross-dataset check (camera location $L_1$, within-camera protocol). (a) Held-out normal frame used as a visual reference. (b) Fully obstructed lens, correctly flagged. (c) Defocused frame, not flagged. (d) Rotated camera, not flagged. Source: Attarha, Shanmugi, and F\"orster (arXiv:2602.05706).}
\label{fig:iot_qualitative}
\end{figure}

Figure~\ref{fig:iot_qualitative} shows representative frames behind these numbers: a held-out normal reference, a correctly flagged full-lens obstruction, a defocused frame the predicate misses, and a rotated-camera frame the predicate misses. The obstruction case is a near-uniform frame with collapsed gradient. The defocus case retains enough residual luminance structure that it falls outside the required mean-shift interval even though a human viewer immediately sees the impairment. The rotation case is nearly indistinguishable from the normal reference in global mean and sampled gradient, since the predicate has no registration cue.

\begin{table*}[t]
\caption{Full independent public image cross-dataset check. $L_1$ and $L_2$ denote independent camera locations; $L_1\!\to\!L_2$ builds the reference from $L_1$ and evaluates on $L_2$ without recalibration. Class columns report per-category alarm rates; \emph{Noimage} is the dataset's name for the obstructed-lens class.}
\label{tab:public_iot_full_appendix}
\centering
\scriptsize
\setlength{\tabcolsep}{3.2pt}
\begin{tabular}{llccccccccc}
\toprule
Protocol & Method & Prec. & Recall & F1 & BAcc & FPR & Normal & Blur & Noimage & Rotated\\
\midrule
Within $L_1$ & Occlusion predicate & 1.000 & 0.065 & 0.122 & 0.533 & 0.000 & 0.000 & 0.000 & 0.195 & 0.000\\
Within $L_1$ & Gradient collapse & 0.679 & 0.712 & 0.695 & 0.686 & 0.340 & 0.339 & 1.000 & 1.000 & 0.135\\
Within $L_1$ & Frame difference & 0.381 & 0.212 & 0.272 & 0.433 & 0.346 & 0.346 & 0.015 & 0.075 & 0.545\\
Within $L_1$ & Histogram difference & 0.691 & 0.774 & 0.730 & 0.713 & 0.349 & 0.349 & 0.973 & 1.000 & 0.350\\
Within $L_2$ & Occlusion predicate & 1.000 & 0.199 & 0.332 & 0.600 & 0.000 & 0.000 & 0.285 & 0.243 & 0.000\\
Within $L_2$ & Gradient collapse & 0.657 & 0.674 & 0.665 & 0.676 & 0.323 & 0.323 & 0.704 & 1.000 & 0.138\\
Within $L_2$ & Frame difference & 0.526 & 0.648 & 0.581 & 0.557 & 0.533 & 0.533 & 0.369 & 0.896 & 0.707\\
Within $L_2$ & Histogram difference & 0.527 & 0.827 & 0.644 & 0.574 & 0.680 & 0.680 & 0.823 & 1.000 & 0.575\\
$L_1\!\to\!L_2$ & Occlusion predicate & 0.938 & 0.155 & 0.266 & 0.573 & 0.009 & 0.009 & 0.027 & 0.386 & 0.006\\
$L_1\!\to\!L_2$ & Gradient collapse & 0.464 & 0.962 & 0.626 & 0.481 & 1.000 & 1.000 & 1.000 & 1.000 & 0.844\\
$L_1\!\to\!L_2$ & Frame difference & 0.369 & 0.649 & 0.470 & 0.325 & 1.000 & 1.000 & 0.969 & 0.084 & 1.000\\
$L_1\!\to\!L_2$ & Histogram difference & 0.471 & 0.959 & 0.632 & 0.495 & 0.969 & 0.969 & 0.946 & 1.000 & 0.916\\
$L_2\!\to\!L_1$ & Occlusion predicate & 1.000 & 0.140 & 0.246 & 0.570 & 0.000 & 0.000 & 0.142 & 0.278 & 0.000\\
$L_2\!\to\!L_1$ & Gradient collapse & 0.793 & 0.576 & 0.667 & 0.713 & 0.151 & 0.151 & 0.702 & 1.000 & 0.025\\
$L_2\!\to\!L_1$ & Frame difference & 0.484 & 0.937 & 0.638 & 0.469 & 1.000 & 1.000 & 1.000 & 0.862 & 0.950\\
$L_2\!\to\!L_1$ & Histogram difference & 0.500 & 1.000 & 0.667 & 0.500 & 1.000 & 1.000 & 1.000 & 1.000 & 1.000\\
\bottomrule
\end{tabular}
\end{table*}

\section{Statistical Procedure}
For method $A$ and baseline $B$, exact McNemar testing is applied to the discordant correctness counts
\begin{align}
n_{10}&=\sum_i \mathbb{I}[\hat y_i^A=y_i\land\hat y_i^B\ne y_i],\\
n_{01}&=\sum_i \mathbb{I}[\hat y_i^A\ne y_i\land\hat y_i^B=y_i].
\end{align}
The exact two-sided $p$ value uses a $\mathrm{Binomial}(n_{10}+n_{01},0.5)$ model. For the F1 difference, we run 10,000 paired bootstrap resamples over sequence indices and recompute $F1_A-F1_B$. We report the 2.5 and 97.5 percentiles. The same fixed predictions are used for both the all-condition and scope-matched analyses.

\section{Public Long-Negative False-Alarm Audit}
Thresholds are frozen from the existing calibration protocol (no tuning on these recordings). VIRAT ground-camera videos are a staged-activity collection whose published protocol contains no camera tampering, supporting the ground-truth-negative justification; the four audited CDnet2014 categories (camera jitter, bad weather, PTZ, turbulence) are hard nuisance-motion conditions with full per-frame change-detection annotations confirming no tamper events. Table~\ref{tab:negative_audit} reports false alarms per camera-hour (FA/h) by method. The full state machine and mean jump are the only methods with zero false alarms on both sources; gradient collapse, frame difference, and histogram difference all fire substantially more often on CDnet2014's harder nuisance motion than on VIRAT's largely static scenes, with histogram difference reaching 46.1 FA/h. Zero events in 9.09 camera-hours bounds the full state machine's rate below 0.33 false alarms per camera-hour (one-sided 95\% Poisson), i.e., below 7.9 per camera-day; the audited recordings are short (VIRAT averages roughly 94~s per clip), so multi-hour single-camera negative recordings remain future work.

\begin{table}[t]
\caption{Public long-negative false-alarm audit. FSM: full state machine; MJ: mean jump; GC: gradient collapse; FD: frame difference; HD: histogram difference. All values are false alarms per camera-hour.}
\label{tab:negative_audit}
\centering
\scriptsize
\setlength{\tabcolsep}{3.5pt}
\begin{tabular}{lrrrrrrr}
\toprule
Source & Rec. & Cam.-h & FSM & MJ & GC & FD & HD\\
\midrule
VIRAT & 329 & 8.61 & 0.000 & 0.000 & 0.000 & 0.116 & 0.000\\
CDnet2014 & 16 & 0.48 & 0.000 & 0.000 & 2.096 & 41.929 & 46.122\\
\bottomrule
\end{tabular}
\end{table}

\section{Supplementary Operational Diagnostics}
The current summary is computed from completed clip-level records in supplementary operational diagnostics. For each clip, predicted tamper is defined as the presence of at least one emitted tamper event. The reference class in this interim snapshot comes from coarse collection-stage clip categorization. This surrogate-supervision protocol is useful for rapid verification, but not for definitive causal or external-validity claims.

Table~\ref{tab:realclip_detail} separates each operational set by label status and by the only quantities its labels can support. The mixed and large stress sets lack verified true negatives, so they are retained only as completed-clip and alarm-yield summaries rather than precision, recall, or F1 evidence. The broad candidate set is also retained only as an alarm-yield observation: 25 of 4,106 completed clips alarmed, while the historical alarm list is superseded by the separately verified FP/TP candidate pool. The FP/TP candidate set is the only subset with a verified true-negative side; it gives TP=13, FP=12, TN=10, FN=0, precision 0.520, recall 1.000, specificity 0.4545, FPR 0.5455, and F1 0.684.

For the broad candidate set, we also audited non-alarming clips to probe false negatives. Four heuristics (including an edge-density terminal-texture screen) screened all 4,081 non-alarming clips, and 69 flagged or sampled clips were manually reviewed. The audit confirmed 2 false negatives and 67 true negatives. Both missed events were close-range covers that collapsed texture while failing the current mean-shift interval, which identifies a concrete predicate weakness rather than a new recall estimate for the whole set. This audit is non-exhaustive and is therefore reported as failure-mode evidence only.

Both confirmed false negatives share the same predicate-level explanation. The close object collapses texture, but the resulting mean shift falls outside the bounded interval of the occlusion predicate in Eq.~\eqref{eq:supp-occlusion}, especially the upper limit $\Delta_{\max}$. Increasing $\Delta_{\max}$ would recover some of these events but also risks converting lighting and exposure changes into occlusion alarms. A safer remedy is to add a color-agnostic terminal texture cue, such as an edge-density or high-frequency-energy floor, and require it to agree with the low-gradient state before extending the accepted mean-shift interval. This is the motivation for the edge-density screening used in the manual FN audit; it is not yet part of the production predicate.

\begin{table}[!t]
\caption{Supplementary Operational Diagnostics}
\label{tab:realclip_detail}
\centering
\scriptsize
\setlength{\tabcolsep}{2.0pt}
\resizebox{\columnwidth}{!}{%
\begin{tabular}{lcccc}
\toprule
Metric & Mixed & Large & Broad cand.\textsuperscript{*} & FP/TP cand.\\
\midrule
Completed clips & 117 & 1,394 & 4,106 & 35\\
Execution failures & 0 & 0 & 9 & 0\\
Verified TN available & no & no & no & yes\\
Reported alarm outcomes & 5 likely TP / 2 likely FP & 5 likely TP / 4 likely FP & label under revision & 13 TP / 12 FP\\
Unverified non-alarm outcomes & 110 & 1,385 & 4,081 & ---\\
Alarms / alarm rate & 7 / 0.060 & 9 / 0.0065 & 25 / 0.0061 & 25 / 0.714\\
Supported accuracy metrics & none & none & none & precision/recall/F1/FPR\\
Verified-set precision & --- & --- & --- & 0.520\\
Verified-set recall & --- & --- & --- & 1.000\\
Verified-set specificity / FPR & --- & --- & --- & 0.4545 / 0.5455\\
Verified-set F1 & --- & --- & --- & 0.684\\
\bottomrule
\end{tabular}
}

\vspace{2pt}
\footnotesize\textsuperscript{*}The broad candidate set's whole-directory positive label was found unreliable after this snapshot; only the raw alarm count is retained until a corrected per-clip label is available. The 25 alarms trace to the same clips as a subset of the FP/TP candidate set (16 confirmed-negative, 9 confirmed-positive there), superseded by that set's fresher complete rerun. A separate, non-exhaustive audit of 69 of its 4,081 non-alarming clips confirmed 2 false negatives and 67 true negatives.
\end{table}

\begin{table}[!t]
\caption{Unlabeled Operational Sweep}
\label{tab:op18k_supp}
\centering
\setlength{\tabcolsep}{4.0pt}
\begin{tabular}{lccc}
\toprule
Set & Completed & Alarms & Alarm rate\\
\midrule
Operational sweep & 4,140 & 20 & 0.0048\\
\bottomrule
\end{tabular}
\end{table}

Table~\ref{tab:op18k_supp} reports a separate unlabeled operational sweep. Records that did not complete are excluded from the denominator and are not treated as negatives, positives, or completed clips. Because the sweep has not been independently annotated, it is reported only as an alarm-yield audit and is not converted into TP, TN, FP, or FN counts. A timestamp audit of the same simulator batch spans 12.42 hours, corresponding to 333 completed clips per hour, with median and 95th-percentile inter-record intervals of 11 and 14 seconds. This is a batch-throughput audit rather than a per-frame latency or energy measurement.